\documentclass[10pt,twocolumn,letterpaper]{article}

\usepackage{cvpr}
\usepackage{times}
\usepackage{epsfig}
\usepackage{graphicx}
\usepackage{amsmath}
\usepackage{amssymb}

\usepackage[breaklinks=true,bookmarks=false]{hyperref}

\cvprfinalcopy % *** Uncomment this line for the final submission

\def\cvprPaperID{****} % *** Enter the CVPR Paper ID here

\begin{document}

%%%%%%%%% TITLE
\title{Class Activation Map Generation by Representative Class Selection and Multi-Layer Feature Fusion}

\author{Fanman Meng, Kaixu Huang, Hongliang Li, Qingbo Wu\\
School of Information and Communication Engineering\\
University of Electronic
Science and Technology of China, Cheng Du, China, 611731\\
{\tt\small fmmeng@uestc.edu.cn}
\and
}

\maketitle
%\thispagestyle{empty}

%%%%%%%%% ABSTRACT
\begin{abstract}
   Existing method generates class activation map (CAM) by a set of fixed classes (i.e., using all the classes), while the discriminative cues between class pairs are not considered. Note that activation maps by considering different class pair are complementary, and therefore can provide more discriminative cues to overcome the shortcoming of the existing CAM generation that the highlighted regions are usually local part regions rather than global object regions due to the lack of object cues. In this paper, we generate CAM by using a few of representative classes, with aim of extracting more discriminative cues by considering each class pair to obtain CAM more globally. The advantages are twofold. Firstly, the representative classes are able to obtain activation regions that are complementary to each other, and therefore leads to generating activation map more accurately. Secondly, we only need to consider a small number of representative classes, making the CAM generation suitable for small networks. We propose a clustering based method to select the representative classes. Multiple binary classification models rather than a multiple class classification model are used to generate the CAM. Moreover, we propose a multi-layer fusion based CAM generation method to simultaneously combine high-level semantic features and low-level detail features. We validate the proposed method on the PASCAL VOC and COCO database in terms of segmentation groundtruth. Various networks such as classical network (Resnet-50, Resent-101 and Resnet-152) and small network (VGG-19, Resnet-18 and Mobilenet) are considered. Experimental results show that the proposed method improves the CAM generation obviously.
\end{abstract}

%%%%%%%%% BODY TEXT
\section{Introduction}
Generating class activation map (CAM) for deep classification network is an important task in computer vision. It is the basis of many weakly supervised computer vision tasks, such as segmentation\cite{li2018weakly,zhou2018weakly,durand2017wildcat,hu2018associating}, fine-grained classification \cite{rodriguez2018attend, sun2018multi,woo2018cbam}, domain adaption \cite{kang2018deep} and detection\cite{gan2018geometry}.

The existing CAM generation method consists of two steps: the training of the classification model, and the generation of the activation map. The training step is the basic step of CAM generation that can be summarized as
\begin{equation}
C = CNN_c (c_i|c_r)
\end{equation}
where $c_i$ is the $i$th class for CAM generation, $c_r$ is the additional classes for comparison. $CNN_c$ is the deep classification network, and $C$ is the classification model by the training. The CAM generation indeed compares the differences between classes $c_i$ and $c_r$ so as to capture the discriminative regions inter the classes to form the activation region, where the activation regions are related to the class set $c_r$ used for comparing.

However, the existing method specifies $c_r$ as all the given classes, i.e., setting $c_r$ as a fixed set $\{c_1, ..., c_n\}$ (where $n$ is the number of the classes), and ignores the using of different classes of $c_r$, which leads to two problems. First, the highlighted region should be distinguished from all classes, resulting in the local region generation. Second, existing methods ignore the impact of the comparison classes on the CAM extraction.
Note that by selecting appropriate $c_r$, more distinct regions can be activated, which results in more accurate activation map generation.

\begin{figure}[t]
\begin{center}
   \includegraphics[width=0.9\linewidth]{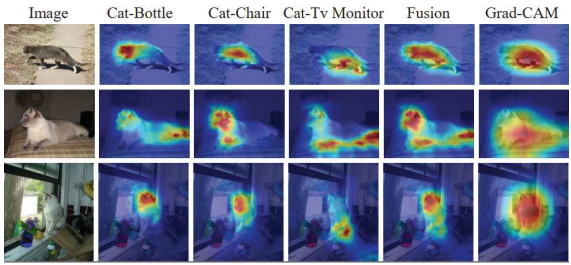}
\end{center}
   \caption{Examples of activation maps highlighted by different classes. The fusion maps and the activation maps by Grad-CAM \cite{selvaraju2017grad} method are also displayed for comparison. }\label{different_introduction}
\end{figure}

In order to observe the affection of the comparison classes, we conduct an experiment on CAM extraction by different comparison classes. Specifically, given the images of a class such as ``Cat'', we consider the binary classification with comparison classes such as ``Bottle'', ``Chair'', and ``Monitor'', and obtain the activation map for each binary classification task. Grad-CAM method \cite{selvaraju2017grad} is used to generate the activation map. Resnet-152 network is employed as backbone network, and PASCAL VOC 2012 dataset is considered. The original images, the corresponding class activation maps by each comparison class, and the activation map by all class (original Grad-CAM method) are displayed in Fig. \ref{different_introduction}.

\begin{table*}
%\tiny
\caption{The results by Grad-CAM \cite{selvaraju2017grad} on PASCAL VOC 2012 dataset, where the binary classification of all class pairs are considered. The value is the mIoU of the CAM (threshold by 0.15) and the segmentation groundtruth. The Top$k$ means to generate CAM by averaging the class activation maps with the best top $k$ classes. ``All'' is the results by fusing CAM of all rest 19 classes. Grad-CAM is the baseline results by Grad-CAM. It is seen that the mIoU values of Top$k$ method are obviously larger than the traditional Grad-CAM method that uses all classes and one multi-class classification model.}
\begin{center}
\resizebox{\textwidth}{30mm}{
\begin{tabular}{cccccccccccccccccccccc}
\hline
 Class Name  & aeroplane & bicycle & bird & boat & bottle & bus & car & cat & chair & cow & diningtable & dog & horse & motorbike & person & pottedplant & sheep & sofa & train & tvmonitor & avg \\
\hline
aeroplane & - & 0.1680 & 0.2390 & 0.1915 & 0.2255 & 0.5525 & 0.3352 & 0.4462 & 0.1155 & 0.3651 & 0.2498 & 0.4839 & 0.4396 & 0.3707 & 0.2960 & 0.1376 & 0.4108 & 0.2295 & 0.4558 & 0.1541 & - \\
\hline
bicycle & 0.2321 & - & 0.2080 & 0.2014 & 0.2571 & 0.5005 & 0.3279 & 0.4945 & 0.1718 & 0.4604 & 0.3047 & 0.4596 & 0.4268 & 0.4759 & 0.2363 & 0.1692 & 0.3833 & 0.3397 & 0.4347 & 0.1962 &  - \\
\hline
bird & 0.3397 & 0.1740 & - & 0.2507 & 0.2641 & 0.5036 & 0.3534 & 0.4079 & 0.1632 & 0.4744 & 0.2443 & 0.4246 & 0.3977 & 0.3931 & 0.3042 & 0.1648 & 0.4240 & 0.3033 & 0.4321 & 0.2219 &  - \\
\hline
boat & 0.3270 & 0.1904 & 0.2786 & - & 0.2103 & 0.5160 & 0.2606 & 0.4919 & 0.1290 & 0.4581 & 0.1840 & 0.4502 & 0.4334 & 0.3843 & 0.3210 & 0.1264 & 0.2826 & 0.2338 & 0.4659 & 0.2399 & -  \\
\hline
bottle & 0.2349 & 0.1813 & 0.2306 & 0.2071 & - & 0.5100 & 0.3354 & 0.5318 & 0.1678 & 0.4359 & 0.2243 & 0.5077 & 0.4053 & 0.5181 & 0.2685 & 0.1997 & 0.3578 & 0.3161 & 0.4250 & 0.2880 &  - \\
\hline
bus & 0.2807 & 0.1862 & 0.2511 & 0.2166 & 0.1919 & - & 0.3300 & 0.4688 & 0.1666 & 0.3859 & 0.2971 & 0.4363 & 0.4457 & 0.3635 & 0.3250 & 0.1511 & 0.3121 & 0.2572 & 0.3617 & 0.1979 &  - \\
\hline
car & 0.1996 & 0.2072 & 0.2216 & 0.2166 & 0.2064 & 0.5539 & - & 0.4519 & 0.1717 & 0.4417 & 0.2570 & 0.4109 & 0.3975 & 0.4631 & 0.2960 & 0.1711 & 0.3494 & 0.2979 & 0.4513 & 0.2547 &  - \\
\hline
cat & 0.1816 & 0.1666 & 0.2306 & 0.2141 & 0.2677 & 0.5412 & 0.2715 & - & 0.1501 & 0.3908 & 0.2834 & 0.3997 & 0.3712 & 0.3833 & 0.2861 & 0.1421 & 0.2845 & 0.2923 & 0.4938 & 0.2167 & - \\
\hline
chair & 0.2796 & 0.2391 & 0.2247 & 0.1981 & 0.4022 & 0.5453 & 0.3700 & 0.5408 & - & 0.3576 & 0.3113 & 0.4728 & 0.4773 & 0.4305 & 0.3087 & 0.2437 & 0.3491 & 0.4226 & 0.4817 & 0.3153 &  - \\
\hline
cow & 0.2669 & 0.1619 & 0.2510 & 0.1858 & 0.2196 & 0.521 & 0.3675 & 0.3717 & 0.1608 & - & 0.2854 & 0.3913 & 0.3807 & 0.4941 & 0.2110 & 0.1528 & 0.3336 & 0.2538 & 0.4354 & 0.1691 & -  \\
\hline
diningtable & 0.2693 & 0.1705 & 0.2273 & 0.1345 & 0.2240 & 0.5389 & 0.2833 & 0.5184 & 0.1766 & 0.3870 & - & 0.4802 & 0.3924 & 0.3474 & 0.2587 & 0.2239 & 0.2960 & 0.3852 & 0.3555 & 0.3292 &  -\\
\hline
dog & 0.2325 & 0.2458 & 0.2242 & 0.2023 & 0.2876 & 0.5457 & 0.3311 & 0.4520 & 0.2111 & 0.3155 & 0.3829 & - & 0.3797 & 0.4956 & 0.3494 & 0.1801 & 0.3269 & 0.3289 & 0.4480 & 0.2963 &  - \\
\hline
horse & 0.2044 & 0.1732 & 0.2131 & 0.2246 & 0.1865 & 0.5400 & 0.3580 & 0.3745 & 0.1529 & 0.3381 & 0.3338 & 0.3984 & - & 0.3935 & 0.3095 & 0.1463 & 0.3631 & 0.2579 & 0.3899 & 0.2116 & -  \\
\hline
motorbike & 0.2742 & 0.1576 & 0.1982 & 0.2009 & 0.2068 & 0.4881 & 0.3456 & 0.4879 & 0.1725 & 0.4218 & 0.2882 & 0.4668 & 0.4103 & - & 0.2919 & 0.1451 & 0.3654 & 0.2736 & 0.3370 & 0.1760 &  - \\
\hline
person & 0.2743 & 0.2003 & 0.2276 & 0.2099 & 0.3940 & 0.5181 & 0.4040 & 0.5078 & 0.1732 & 0.4369 & 0.2992 & 0.4903 & 0.4432 & 0.4281 & - & 0.2120 & 0.3635 & 0.3702 & 0.4201 & 0.3189 & -  \\
\hline
pottedplant & 0.2202 & 0.2072 & 0.2295 & 0.2064 & 0.2654 & 0.4895 & 0.3435 & 0.4209 & 0.2207 & 0.3458 & 0.3293 & 0.4130 & 0.3768 & 0.4251 & 0.2849 & - & 0.3394 & 0.3397 & 0.4114 & 0.2802 &  - \\
\hline
sheep & 0.2551 & 0.1818 & 0.2288 & 0.2934 & 0.2336 & 0.5306 & 0.3292 & 0.3953 & 0.1161 & 0.3892 & 0.2630 & 0.3801 & 0.3399 & 0.4582 & 0.3062 & 0.1625 & - & 0.2553 & 0.4877 & 0.2057 &   -\\
\hline
sofa & 0.2652 & 0.1330 & 0.1765 & 0.1719 & 0.2500 & 0.4075 & 0.2841 & 0.4846 & 0.1572 & 0.3315 & 0.3161 & 0.3950 & 0.3409 & 0.4991 & 0.2147 & 0.2157 & 0.2995 & - & 0.3732 & 0.3990 & -  \\
\hline
train & 0.2273 & 0.1769 & 0.2140 & 0.1742 & 0.2136 & 0.4252 & 0.2729 & 0.4303 & 0.1436 & 0.3634 & 0.3310 & 0.4104 & 0.3653 & 0.3959 & 0.2819 & 0.1336 & 0.3409 & 0.2494 & - & 0.2621 & -  \\
\hline
tvmonitor & 0.1991 & 0.1377 & 0.1872 & 0.1764 & 0.2349 & 0.5241 & 0.3091 & 0.5383 & 0.1714 & 0.3147 & 0.3698 & 0.4246 & 0.2930 & 0.3724 & 0.2562 & 0.1809 & 0.3185 & 0.3391 & 0.4287 & - & - \\
\hline\\
\hline
Top1 & 0.3397 & 0.2458 & 0.2786 & 0.2934 & 0.4022 & 0.5539 & 0.404 & 0.5408 & 0.2207 & 0.4744 & 0.3829 & 0.5077 & 0.4773 & 0.5181 & 0.3494 & 0.2437 & 0.424 & 0.4226 & 0.4938 & 0.399 & 0.3986 \\
\hline
Top2 & 0.3397 & 0.2433 & 0.2781 & 0.2756 & 0.3641 & 0.5923 & 0.3861 & 0.5617 & 0.2253 & 0.4958 & 0.4152 & 0.5203 & 0.4788 & 0.5401 & 0.3677 & 0.2256 & 0.4251 & 0.4105 & 0.5165 & 0.3672 & 0.4015 \\
\hline
Top3 & 0.324 & 0.2429 & 0.2709 & 0.2685 & 0.3218 & 0.6159 & 0.3967 & 0.5685 & 0.2242 & 0.5068 & 0.4107 & 0.5193 & 0.483 & 0.554 & 0.3749 & 0.2253 & 0.4254 & 0.4096 & 0.5257 & 0.3501 & 0.4009 \\
\hline
Top4 & 0.3165 & 0.2379 & 0.2686 & 0.2614 & 0.3046 & 0.6203 & 0.4029 & 0.5801 & 0.2215 & 0.5019 & 0.4061 & 0.5249 & 0.4789 & 0.5515 & 0.379 & 0.221 & 0.4244 & 0.409 & 0.5301 & 0.3408 & 0.3991 \\
\hline
All & 0.2703 & 0.1984 & 0.2326 & 0.2133 & 0.2513 & 0.5991 & 0.3704 & 0.5244 & 0.1745 & 0.4631 & 0.3235 & 0.4998 & 0.4264 & 0.497 & 0.3715 & 0.1885 & 0.3874 & 0.3212 & 0.4817 & 0.2491 & 0.3522 \\
Grad-CAM & 0.1941 &  0.1377 &  0.1715  & 0.1576 &  0.2194 &  0.4469 &  0.2892 &  0.4491  & 0.1533 &  0.3778 &  0.2249 &  0.4053 &  0.3046 &  0.3354 &  0.3452 &  0.1765 &  0.2726 &  0.2244 &  0.3483 &  0.2138 &  0.2724\\
\hline
\end{tabular}}
\end{center}
\label{bianry_test}
\end{table*}

From the results, we can summarize that:
\begin{itemize}
\item The CAM extracted by different comparison classes are different. For example, the CAM by ``Bottle'' is very different from the one by ``Monitor''.
\item The CAM generated by all classes is blurred, while the results generated by the binary classification are more clear.
\item The CAMs generated by these classes are complementary, such as the maps by ``Chair'' and ``Monitor''.
\end{itemize}

Table \ref{bianry_test} shows the objective values of the activation maps generated by all the binary classification models in PASCAL VOC validation dataset that further supports our observations. The activation map is extracted by Grad-CAM \cite{selvaraju2017grad}. The objective value is the mIoU of the CAM mask (by threshold 0.15) and the segmentation groundtruth. The Top-$k$ means to generate CAM by averaging the class activation maps of the best top $k$ classes. ``All'' is the results by fusing CAM of the rest 19 classes. ``Grad-CAM'' is the baseline results by Grad-CAM. It is seen that the mIoU values of Top-$k$ method are obviously larger than Grad-CAM method \cite{selvaraju2017grad} that uses all classes and multi-class classification model (0.4009 vs 0.2724). Hence, selecting complementary classes are useful to improve the CAM generation. It is also seen that using top three classes has been able to achieve significant improvement. It means that when we are able to choose a small number of proper classes so that their activation regions are complementary, the CAM generation can be effectively enhanced.

Based on such motivation, we propose a CAM generation method based on class selection strategy. On the one hand, the complementary classes are selected to achieve more comprehensive generation of the activation maps. On the other hand, a few of classes rather than all classes are considered so that small CNN-based network can be used.
The proposed method consists of two steps. In the first step, we select the representative classes by clustering strategy. To this end, we use false positive of the classification model to represent the class relationship and generate the class similarity matrix. The improved k-means based clustering method is then proposed to divide the classes into several clusters.
Finally, we select representative classes from different clusters to guarantee their differences.

In the second step, we generate the CAM by the representative classes. Instead of constructing a multi-classification model as used in the existing methods, we use multiple binary classification models. Specifically, given a class $c_i$ and its representative classes $s=\{s_1,\cdots, s_N\}$, we generate the activation maps by each binary classification task $(c_i,s_k)$, and then combine them to generate the class activation map. We propose a multi-layer analysis fusion method with the aim of combining low-level and high-level features to enhance the CAM generation.

We conduct experiments on PASCAL VOC 2012 dataset with segmentation groundtruth. Several backbone networks including classical CNN network and small CNN network are considered. The experimental results show that the proposed method can obviously improve the CAM extraction with larger mIoU values.

\section{Related Work}

Several class activation map generation methods have been proposed recently \cite{Fong2018Interpretable,zhang2017interpretable,park2018multimodal,chen2018attention,owens2018audio}. Some methods first introduce the local mechanisms such as global max pooling \cite{Oquab2015Is} and log-sum-exp pooling \cite{ pinheiro2015image} to highlight the activation regions. Then, Zhou et al. \cite{zhou2016learning} propose CAM method that uses average pooling layer and FC layer to modify the high layers of the classification network. The CAM is then highlighted by averaging the last convolution layer weighted by the weights of the following FC layer. Better CAM can be generated. After that, several methods are proposed to enhance the CAM generation. An important extension is the Grad-CAM \cite{selvaraju2017grad} that uses the gradient signals to form the weights directly. Hence, the drawbacks of CAM that modifies the initial network is avoided.

By seeing the fact that the CAM generated by \cite{zhou2016learning} are usually local part region rather than the object regions, researchers try to solve such drawbacks to generate more activation regions. Erasing strategy that finds more activation regions by erasing the obtained regions is a useful solution, and is widely used. For example,
Wei et al. \cite{Wei2017Object} propose an adversarial erasing method that erases the obtained activation regions, and implement the CAM extraction again to obtain the next most activation regions. Such process iteratively implemented to find more regions.
Kim et al. \cite{Kim2017Two} propose a two-phase based method that first extract the activation map by CAM method. Then, the activation region is erased, and the rest region is send to the second phase learning by element-wise multiplication to extract more activation regions.
Li et al. \cite{li2018tell} design additional loss for activation map that makes the output of classification network as small as possible when the activation regions are erased.
Recently, Zhang et al. \cite{zhang2018adversarial} propose the adversarial complementary learning (ACoL) method that uses two classifier to extract the complementary object regions through the erasing operation. Better results can be obtained. However, these methods consider all classes, and fail to analyze the affects of the selection of the comparison classes that can provide more complementary regions.

\section{The Proposed Method}
\begin{figure*}[t]
\begin{center}
   \includegraphics[width=0.85\linewidth]{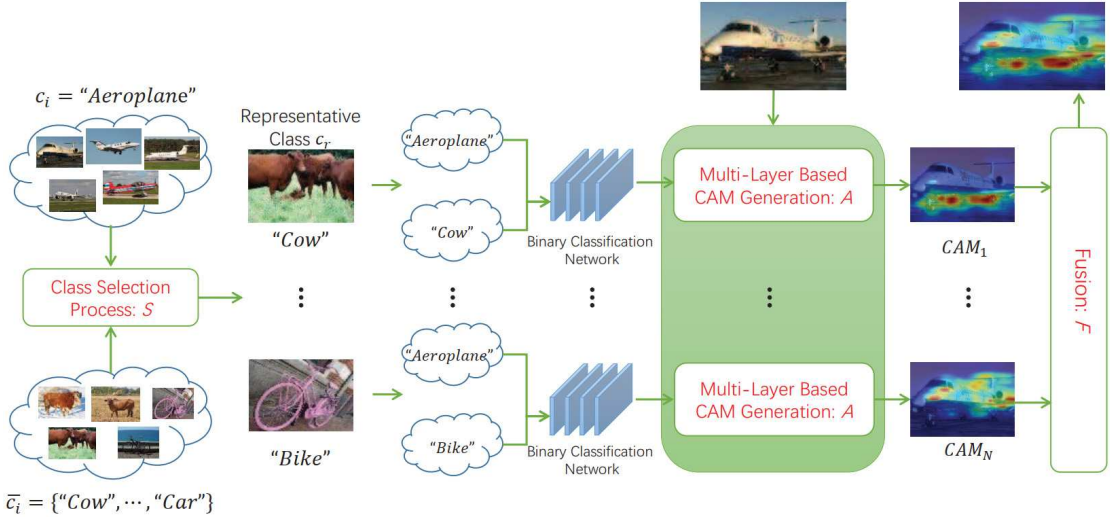}
\end{center}
   \caption{The flowchart of the proposed method. Given a class such as ``Aeroplane'', we first select the representative class $c_r$ by class selection process $S$. Then, for each representative class, we construct binary classification, and use multi-layer CAM generation $A$ to obtain activation map. Finally, we fuse the activation maps of all representative classes to obtain our classification map.}\label{framework}
\end{figure*}

The framework is shown in Fig. \ref{framework}, where our method consists of three steps such as class selection step $S$, CAM generation step $A$, and fusion step $F$.

\textit{Class selection step $S$}: Given a class $c_i$ with images $I_i = \{I_{i1},\cdots, I_{in_i}\}$, and the rest classes $\bar{c}_i=\{c_1,\cdots,  c_{i-1}, c_{i+1}, \cdots, c_n\}$, we first select $N$ classes from $\bar{c}_i$ and build the representative class $s = \{s_1, \cdots, s_N\}$ that will be used for extraction. We denote such process as $s=S(c_i, \bar{c}_i)$.

\textit{CAM generation step $A$}: For each class $s_k \in s, k=1, ..., N$, we construct a binary classification task by class pair $(c_i, s_k)$, and extract the activation map $M_{ij}^k$ for each image $I_{ij}$ based on the binary classification by CAM extraction process $A$, i.e., $M_{ij}^k = A(I_{ij}, s_k)$.

\textit{Fusion step $F$}: Given the class activation maps $M_{ij}^k, k=1, \cdots, N$ for image $I_{ij}$, we combine these activation maps by $F$ process, i.e., $M_{ij} = F(M_{ij}^k),k=1, \cdots, N$, and the final class activation map $M_{ij}$ is generated.

Here, we propose a clustering based method for process $S$, and multiple binary based CAM fusion framework and a multi-layer based CAM extraction method for process $A$. The merging method $F$ is achieved simply by the map averaging. We next detail these steps.

\subsection{The Class Selection Step: $S$}

Intuitively, the class in $s$ can be selected randomly. However, such process ignores the differences of classes on activating the regions. Here, we select the regions based on class similarity, and try to select classes that are very dissimilar. Two selection methods are proposed.

\subsubsection{Selection by Similarity Sorting}
The first one is a similarity-based sorting and selection method. Specifically, the similarity is represented by a matrix $B$ where $B_{ij}$ depicts the similarity between classes $c_i$ and $c_j$. For the class $c_i$, the $i$th row means the similarities between the rest classes and $c_i$. The larger the value, the more similarity the two classes. By the similarity, we sort the class index by descending order.
The classes in different positions has different similarity to $c_i$, and they are different with each other. Therefore, we select classes from different positions. For simplicity, we use the fixed position for the selection.

\begin{figure}[t]
\begin{center}
   \includegraphics[width=0.8\linewidth]{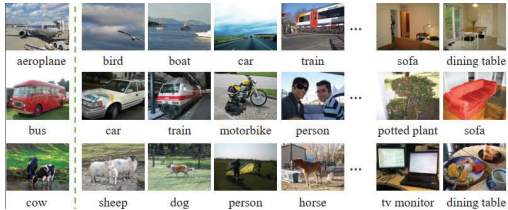}
\end{center}
   \caption{The ranked classes of the class ``aeroplane'', ``bus'', and  ``cow'' by the proposed method.}\label{sort}
\end{figure}

An example can be found in Fig. \ref{sort}, where the ranked classes of the class ``aeroplane'', ``bus'', and  ``cow'' are displayed. It is seen that the ``Car'', and ``Train'' are similar to the class ``Bus''. They are highly likely to have the same activation region. Therefore, we intend to select one of them only as our representative class.

\subsubsection{Generating the Similarity Matrix by False Classification}
The sorting based selection method depends on accurate similarity matrix $B$. We use the false positive probability of the classification model to construct the similarity matrix. Specifically, we first train the classification model for all classes. Given an image $I_{ik}$, the output of the final softmax layer is the probability vector $v_{ik}=\{v_{ik}(1),\cdots,v_{ik}(n)\}$ where $v_{ik}(l)$ means the probability of the image classifying to class $c_l$. Then, for a pair of classes $(c_i,c_j)$, we sum the probability value of the images in $c_i$ classifying to class $c_j$, and use the sum value as the similarity value $B_{ij}$, i.e.,
\begin{equation}
B_{ij} = \sum_{k} v_{ik}(j)
\end{equation}
It is seen that the larger the value of $B_{ij}$, the more similar the two classes $c_i$ and $c_j$.

%It is worthy to note that a relationship graph can be generated. Fig. shows the relationship graph of PASCAL VOC dataset

\subsubsection{Selection by Clustering Strategy}

For the sorting-based selection method, it ensures that classes similar to $c_i$ have similar relationships, such as ``Cow'' and ``Dog'' to ``Cat''. However, the relationships between its dissimilar classes are still unknown. To this end, we further consider all the relationships between classes, and use clustering to enhance the class selection.

Our idea is simple that we cluster all the classes into $N$ clusters according to their similarity relationships. Therefore, the similar classes are grouped into one class. We then select the class from different clusters, and form the representative class set.

The clustering is based on the similarity matrix $B$. Here, we use the method above to construct the similarity matrix $B$. Since the similarity is obtained by the classification model, it is hard to measure the similarity between a sample and a cluster center. The traditional k-means method cannot be directly applied here. To this end, we propose an improved K-means method by using the averaging distance to replace the distance between samples and cluster center.

Specifically, we first divide all the classes into $N$ clusters randomly. Then, for each class $c_i$, we calculate the average similarity value between $c_i$ and the samples in the cluster $g_j$, and move $c_i$ into the cluster with the largest average similarities. The above processes are iterative implemented until the stop of the updating. After clustering, we select one class from each cluster to form $c_s$. Here, we sort the classes in each cluster $g_j$ by the similarity values with the given class. Then, we select the $k'$th class from each cluster as $s_j$.

\begin{figure*}[t]
\begin{center}
   \includegraphics[width=0.8\linewidth]{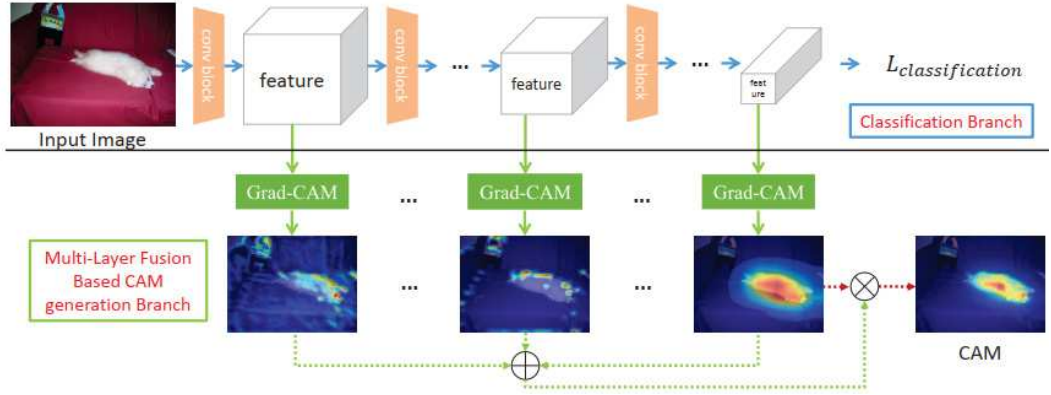}
\end{center}
   \caption{The multi-layer based class activation map generation.}\label{cam_improved}
\end{figure*}
\subsection{The Class Activation Map Generation Step: $A$}
Rather than training a multi-class based model, our CAM extraction is based on multiple binary classification models. Specifically,
given a class $c_i$ and a set of selected class $s=\{s_1,\cdots,s_N\}$, we generate the activation map of $I_{ik}\in c_i$ by considering each class in $s$. First, we generate the activation map $M_{ij}^k$ based on the binary classification model with each class pair $(c_i,s_k)$. Then, we fuse the activation maps $M_{ij}^k, k=1,\cdots, N$ to produce the final activation map.

Given two classes $(c_i, s_k)$, we extract the CAM based on Grad CAM extraction method. Since the Grad-CAM uses high-level convolution features only, the high-level features lack location details, and makes the activation map blurred. Here, we improve grad-cam by multi-layer fusion. Specifically, we first extract the activation map by different feature layers. Then, we add these activation maps to obtain the final activation map.

The proposed network is shown in Fig. \ref{cam_improved}. Compared to the traditional Grad-CAM method, we not only generate the activation map in the last layer, but also output the activation map in the front layer, i.e., we add the grad-cam generation branch to each layer and generate a series of class activation maps, as shown in the figure. The final activation map $M_k$ is then achieved by fusing the previous activation maps.
The fusion manner can be found in Fig. \ref{cam_improved}.

In training the classification network, many classical networks such as VGG-16, Resnet-50, Resnet-101, and Resnet-152 can be selected. Since we use binary classification in generating the CAM, the network with small size such as Resnet-18 and mobileNet can also be employed.

\subsection{The Fusion Step: $F$}

After obtaining the activation map $M_{ij}^k$ for all the classes in $s$, we average these activation maps, and obtain the final activation map, i.e., $M_{ij}=\sum_{k=1}^{N} M_{ij}^k$.

\section{Experiment}
\subsection{Dataset}
We verify our method by PASCAL VOC 2012 and COCO 2017 dataset. The PASCAL VOC 2012 contains 20 classes. The training dataset with 11658 images, and the validation dataset with 1449 images are used.
In our experiment, all the classification models such as 20-class based classification model and multiple binary classification models are trained by the training dataset.  The COCO 2017 dataset consists of 80 classes. We train the classification network on training dataset (by images with single class label only) and extract the class activation map from the validation dataset.

\subsection{Experiment Setup}

In normalizing the images, the short border of all images are resized to 224. Then, the center region with size $224 \time 224$ is cut out from the resized image as the normalized images. In our training, our batch size is set to 24.

Our 20-class based classification network is initialized by model pretrained on ImageNet. We set the initial learning rate to 0.0001. The learning rate is dynamically reduced with scale 0.5 when the reduction of the loss is very small within 5 epochs. A total of 200 epochs are trained in our experiments.

\begin{figure*}[t]
\begin{center}
   \includegraphics[width=0.8\linewidth]{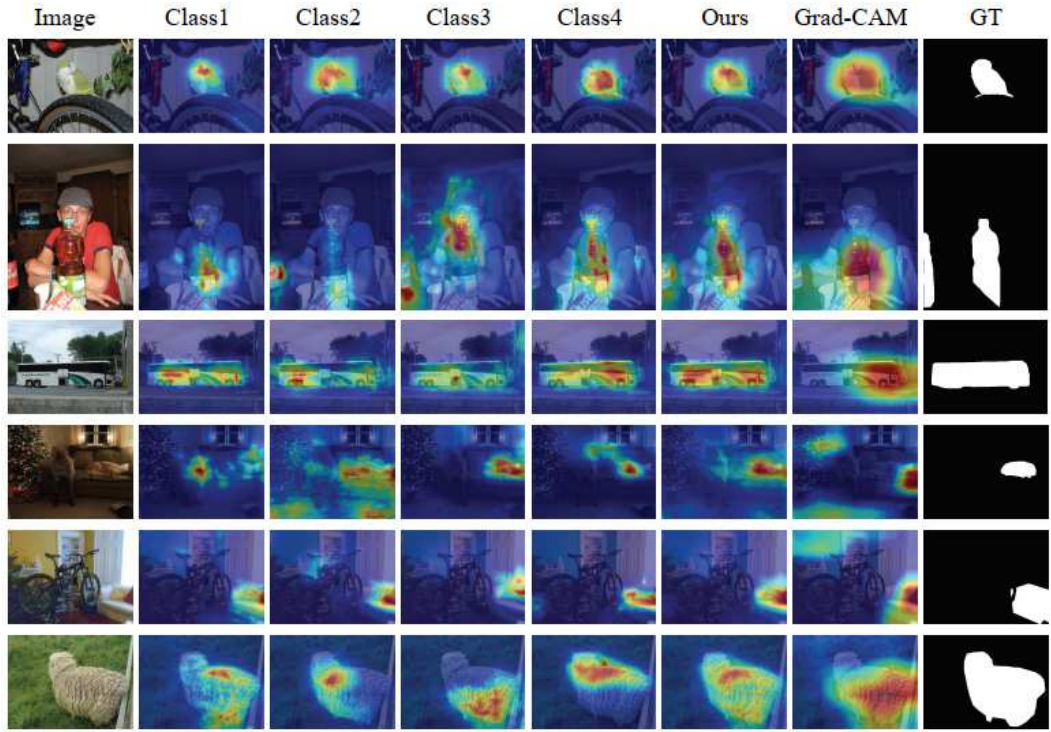}
\end{center}
   \caption{The class activation map generated by the proposed method. The activation map by each representative class, our method, Grad-CAM, and groundtruth are displayed.}\label{cam_res_all}
\end{figure*}

\begin{figure*}[t]
\begin{center}
   \includegraphics[width=0.8\linewidth]{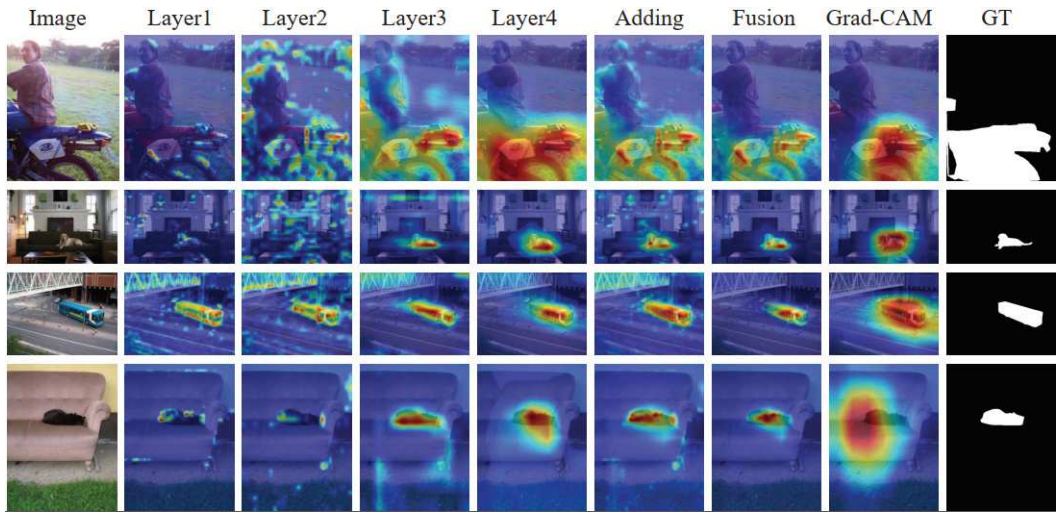}
\end{center}
   \caption{The class activation maps of different layers and their fusion by our multi-layer based CAM generation. The activation maps by Grad-CAM are also displayed for comparison.}\label{cam_res_layer}
\end{figure*}

In training the binary classification network, the network pretrained on ImageNet is used as the initial model. We set the learning rate to 0.0001. we end the training when the classification accuracy is larger than 0.95 or the epoch number is larger than 10. We train our network on the NVIDIA GeForce GTX1080 with memory of 8GB using the PyTorch 0.4 framework.

\subsection{Implementation Details}
In generating the similarity matrix, we use the VGG-19 as the backbone network to train the classification model. To generate accurate classification model, we select the images with single class label for training. Here, we use the images from PASCAL VOC dataset for simple. Note that  the training images from other datasets such as COCO and ImageNet can also be used for the training here, since the classification network is used only to generate the similarity relationships between classes.

We set the cluster number as four empirically to make a trade off between the performance and training burden. We use the clustering method described above to cluster 20 classes into four classes with no less than 4 samples per cluster. It is interesting to see that the final clustering result by our method is \{aeroplane, bird,  cow, sheep\}, \{bicycle, bus, car, motorbike\}, \{boat, bottle, dining table, horse, person, potted plant, train\}, \{cat, chair, dog, sofa, tv monitor\} for PASCAL VOC dataset.

In generating the multi-layer CAM, for Resnet, we generate the CAM of each block (a total of 4 feature blocks), and add them together. Then, we multiply the sum map by the CAM of the final layer to obtain the final CAM. For VGG network, we generate the CAM from the convolution feature before max pooling layers (a total of 5 feature layers). The fusion method is the same to the Resenet. For Mobilenet, we add the CAM generation after each down-sampling layer.

\subsection{Results on PASCAL VOC dataset}
\subsubsection{Subjective Results}

Some class activation maps obtained by our method are displayed in Fig. \ref{cam_res_all}, where the original images, the CAM by all classes (Grad-CAM), the CAM by our selected representative classes, and the final CAM are displayed. The Resnet-101 network is used for training. It is seen that the activation maps by these representative classes are different, and they are complementary. The class activation maps by our fusion method are better than the CAMs of each representative class and the initial Grad-CAM, which demonstrates the effectiveness of the proposed method.

Fig. \ref{cam_res_layer} displays some activation maps by different layers. The original image, the class activation maps of different layers, and the fusion CAM are displayed. Resnet-101 is used.
It is seen that the CAM by low-level layer depicts edges well but contains noise regions. The CAM of the high-level is blurred, but with more accurate localization. By their combination, the final CAM is more accurate.

%Small network

\subsubsection{Objective Results}

We next show the objective results by our method and the comparison method. As we mainly verify the enhancement of our method on the existing class activation generation methods, we compare mainly with the baseline method such as Grad-CAM. Since the CAM generation is used widely in weakly supervised semantic segmentation, we use more detailed segmentation annotations as the groundtruth. The mIoU is used for verification. Larger the value of mIoU, better the CAM extraction. Since the CAM is a probability map rather than binary segmentation task, as usually used in the CAM evaluation, we use a threshold $T=0.15$ to threshold the class activation map, and obtain the segmentation mask.

We display the results in Table \ref{res_miou_validation} and Table \ref{res_miou_train} for PASCAL VOC 2012 validation and training dataset respectively. The mIoU values by the baseline method (Grad-CAM \cite{selvaraju2017grad}), the random selection method (selecting the representative classes randomly), the proposed cluster-based classification method and their combination with the proposed processes such as multi-layer CAM generation $A$ and fusion method $F$ are displayed. It is interesting to see from Table \ref{res_miou_validation} that the random method improves the baseline method on network with small size, such as VGG-19,  and Res-50, while the results becomes worse when the network is large, such as Res-101 and Res-152. By using the proposed multi-layer CAM generation method (i.e., process $A$), the results on all the networks are obviously improved. Such observation is also supported by Table \ref{res_miou_train} for training data. This demonstrates the effectiveness of the proposed multi-layer CAM generation.

Compared the networks in the row of ``Baseline'', it is seen that larger network obtains better mIoU values  such as 29.21 for Res-152 and 27.72 for Res-50.  Compared the proposed method ``S-1+A+F'' with ``Random+F'', we can see our method obviously improves the CAM generation.  Moreover, the best mode for each network on validation network are ``S-4+A+F'', ``S-1+A+F'' and ``S-1+A+F'' for Res-50, Res-101 and Res-152, respectively.  For training dataset, the best mode are ``S-4+A+F'', ``S-1+A+F'', and ``S-1+A+F'' for Res-50, Res-101 and Res-152 that are the same to the validation dataset. This demonstrates the effectiveness of our combination mode.

In the view of ablation, we can see that when we delete our selection strategy ($S$ and $Rank$), and our multi-layer CAM generation $F$, the values decrease very large (34.66 to 27.24 on validation dataset, and 35.93 to 29.41 on training dataset). Meanwhile, when we delete the process $A$, the mIoU values also decrease obviously. But when combing the selection strategy and multi-layer CAM generation, the proposed method becomes better than each of them. This further demonstrates the effectiveness of our combination.

%Moreover, for validation dataset that contains small number of images, the proposed method (we select ``S-1+A+F'') is obviously better than baseline (27.94 vs 34.66). But the value is smaller than ``Sort-a+A+F'' that uses fixed selection manner  and our ``A'' and ``F'' process (34.66 vs 35.87). By varing our selection position, the best value of our method (``S-4+A+F'') is 0.still better than ``Sort-a+A+F''

\begin{table}
\small
\caption{The mIoU values by baseline and the proposed method on PASCAL VOC validation dataset. The mIoU values by baseline and the proposed method. ``Baseline'' is the Grad-CAM since we use it as the basic CAM generation method. ``Random'' is the selection manner that randomly selects the representative classes. ``S'', ``A'', and ``F'' are our cluster based class selection, the multi-layer CAM generation, and our multiple binary classification based CAM generation method. ``Rank-a'' and ``Rank-b'' means the proposed fixed selection orders such as [3,9,14,17] and [3,8,13,18] respectively. ``S-k'' means the kth class ranked by the similarity is selected for each cluster. }
\begin{center}
%\resizebox{\textwidth}{30mm}{
\begin{tabular}{ccccc}
\hline
Method   & Res-50 & Res-101 & Res-152\\ \hline
Baseline \cite{selvaraju2017grad}& 25.49 & 27.94 & 27.24\\ \hline
Random+F  & 27.06 & 26.38 & 25.72\\ \hline
Random+A+F & 28.77 & 32.68 & 33.56\\ \hline
Rank-a +F  & 27.03 & 26.54 & 25.80\\ \hline
Rank-a+A+F  & 29.34 & 32.57 & 33.39\\ \hline
Rank-b +F & 26.64 & 26.16 & 25.37\\ \hline
Rank-b+A+F  & 29.17 & 32.86 & 33.30\\ \hline
S-1+F   & 27.19 & 27.25 & 26.60\\ \hline
S-1+A+F & 28.40 & $\mathbf{33.43}$ & $\mathbf{34.66}$\\ \hline
S-2+F   & 27.25 & 26.72 & 26.14\\ \hline
S-2+A+F  & 28.66 & 32.88 & 33.48\\ \hline
S-3+F  & 26.96 & 26.03 & 25.78\\ \hline
S-3+A+F  & 29.40 & 32.00 & 33.44\\ \hline
S-4+F   & 26.27 & 25.45 & 24.76\\ \hline
S-4+A+F & $\mathbf{29.89}$ & 32.57 & 32.95\\ \hline
\end{tabular}
\end{center}
\label{res_miou_validation}
\end{table}

\begin{table}
\small
\caption{The mIoU values by baseline and the proposed method on PASCAL VOC training dataset. The terms in the first row is the same to the ones in Table \ref{res_miou_validation}}
\begin{center}
%\resizebox{\textwidth}{30mm}{
\begin{tabular}{cccc}
\hline
Method   & Res-50 & Res-101 & Res-152\\ \hline
Baseline \cite{selvaraju2017grad}& 27.72 & 29.41 & 29.21 \\ \hline
random +F&    28.92 & 28.97 & 27.87 \\ \hline
random+A+F & 30.15 & 33.80 & 35.11 \\ \hline
Rank-a+F & 29.15 & 28.88 & 27.86 \\ \hline
Rank-a+A+F & 30.33 & 33.98 & 34.96 \\ \hline
Rank-b+F & 29.04 & 28.51 & 27.57 \\ \hline
Rank-b+A+F & 30.23 & 33.98 & 34.71 \\ \hline
S-1+F & 29.53 & 29.39 & 28.75 \\ \hline
S-1+A+F & 29.72 & $\mathbf{34.43}$ &$\mathbf{35.93}$ \\ \hline
S-2+F & 29.41 & 28.74 & 28.05 \\ \hline
S-2+A+F & 29.76 & 34.26 & 34.75 \\ \hline
S-3+F & 29.08 & 28.35 & 27.94 \\ \hline
S-3+A+F & 30.26 & 33.17 & 34.82 \\ \hline
S-4+F & 28.69 & 27.72 & 26.96 \\ \hline
S-4+A+F & $\mathbf{31.01}$ & 33.68 & 34.22 \\ \hline
\end{tabular}
\end{center}
\label{res_miou_train}
\end{table}

We also show the mIoU values by network with small size, since our method is based on binary classification tasks that can be implemented by small network. The results are displayed in Table \ref{res_samll_network}, where VGG-19 (VGG-19), Resnet-18 (Res-18) and Mobilenet-v2 (Mob) are displayed. It is seen that our method obviously improves the mIoU than baseline method. Meanwhile, sorting-based selection method obtains the best results among all of these manners.  This indicates that sorting-based selection should be employed for small network. However, the mIoU values by cluster-based selection method are very close to the values of the sorting-based method (34.62 vs 35.87, 30.90 vs 31.23, 26.21 vs 26.31 for VGG-19, Res-18, and Mobilenet respectively). This further indicates the effectiveness of the  proposed cluster-based selection method. It is also seen that the mIoU by very small network such as Res-18 and mobilenet are 31.23 and 26.31 by our method, which outperforms or be comparable to the baseline results by deeper network Res-152. We will study the implementation of the proposed method on small network to accelerate the running speed of our method further.

\begin{table}
\small
\caption{The mIoU values by small network, such as VGG-19 (VGG-19), Resnet-18 (Res-18) and mobilenet-v2 (Mobilenet) on validation dataset. The terms in the first row is the same to the ones in Table \ref{res_miou_validation}.}
\begin{center}
%\resizebox{\textwidth}{30mm}{
\begin{tabular}{cccc}
\hline
Method & VGG-19 & Res-18 & Mobilenet \\ \hline
Baseline \cite{selvaraju2017grad} & 22.43 & 19.65 & 25.54  \\ \hline
Random & 32.02 & 24.57 & 26.07  \\ \hline
Random+A+F & 34.54 & 30.83 & 23.03  \\ \hline
Rank-a & 34.62 & 25.19 & $\mathbf{26.31}$  \\ \hline
Rank-a+A+F & $\mathbf{35.87}$ & $\mathbf{31.23}$ & 23.89 \\ \hline
Rank-b & 32.09 & 24.40 & 25.86\\ \hline
Rank-b+A+F & 32.97 & 30.73 & 23.28 \\ \hline
S-1+F & 33.54 & 24.53 & 25.19 \\ \hline
S-1+A+F &34.62 & 29.55 & 22.74 \\ \hline
S-2+F & 32.56 & 24.88 & 25.47 \\ \hline
S-2+A+F &33.37 & 30.90 & 23.00 \\ \hline
S-3+F &  32.85 & 24.61 & 25.76 \\ \hline
S-3+A+F &34.83 & 30.73 & 23.23 \\ \hline
S-4+F & 31.24 & 24.36 & 26.21 \\ \hline
S-4+A+F &34.95 & 30.73 & 23.45 \\ \hline
\end{tabular}
\end{center}
\label{res_samll_network}
\end{table}

\subsection{Results on COCO 2017 Dataset}
We next verify the proposed method on COCO 2017 dataset, which consists of 80 classes. In our verification, the mode ``S-1+A+F'' and ResNet101 are used. We train the classification network on training dataset (by images with single class) and extract the class activation map from the validation dataset.

\begin{figure*}[t]
\begin{center}
   \includegraphics[width=0.95\linewidth]{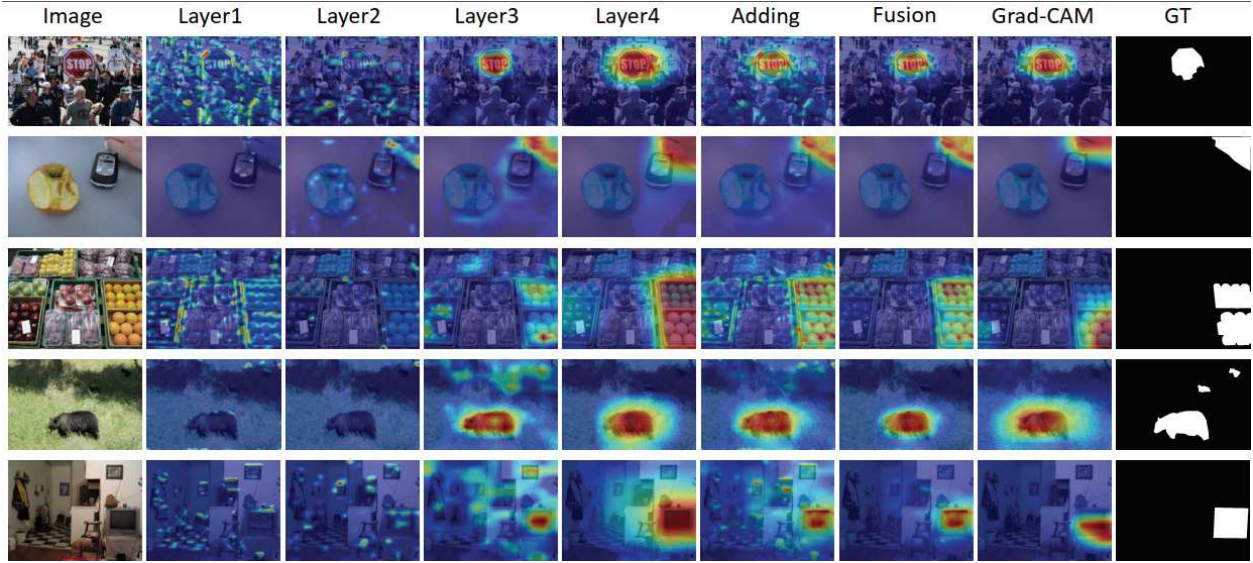}
\end{center}
   \caption{The activation maps by the proposed method on COCO 2017 dataset. ``Layer1'', ``Layer2'', ``Layer3'', ``Layer4'' and ``Adding'' are the CAMs of four different layers, and the sum of the four maps. ``Fusion'' is our final results. ``Grad-CAM'' and ``GT'' are the baseline method \cite{selvaraju2017grad} and groundtruth, respectively.}\label{res_coco_layers}
\end{figure*}

The results by our method on COCO 2017 dataset are displayed in Fig. \ref{res_coco_layers}, where the original images (``Image''), the activation maps of different layers (``Layer1''-``Layer4''), the sum of the activation maps of the layers (``Adding''), and our final activation map (``Fusion'') are displayed. The activation maps by baseline method (``Grad-CAM'') and the groundtruth (``GT'') are also displayed for comparison. It is seen that the proposed method extracts the activation maps successfully from these images. Moreover, the activation maps by the proposed method (``Fusion'') are better than the rest maps and the baseline method \cite{selvaraju2017grad}.

\begin{figure*}[t]
\begin{center}
   \includegraphics[width=0.95\linewidth]{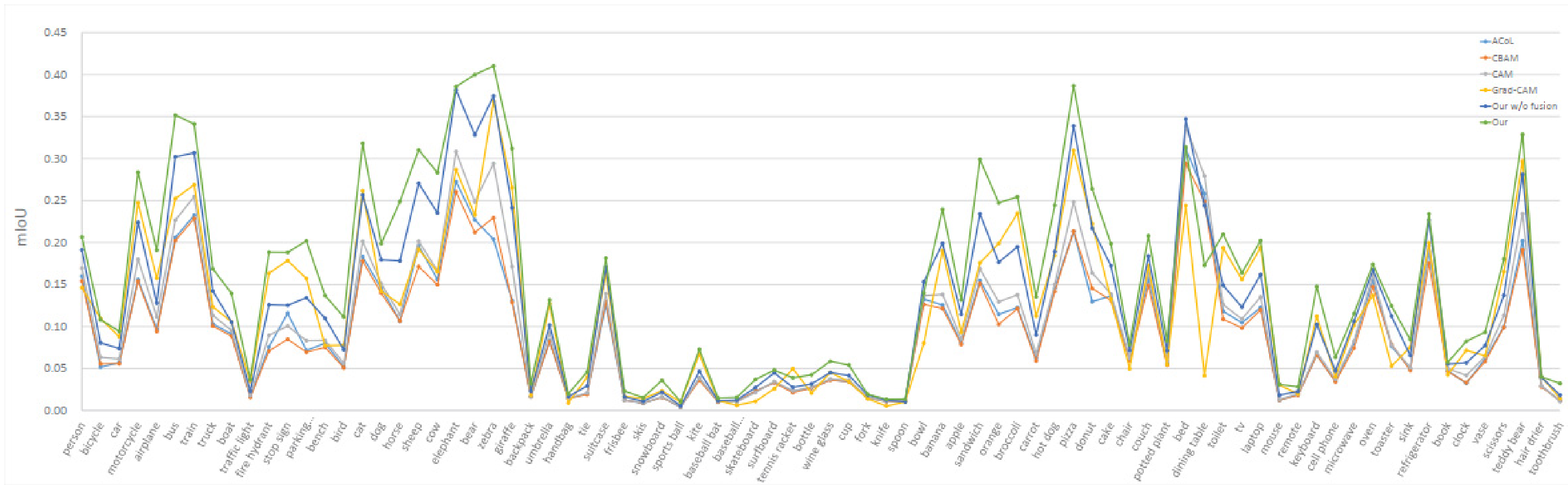}
\end{center}
   \caption{The mIoU values on COCO 2017 dataset by the proposed method and the existing methods. x-axis: the class name. y-axis: the mIoU value.}\label{coco_res2}
\end{figure*}

\begin{figure*}[t]
\begin{center}
   \includegraphics[width=0.95\linewidth]{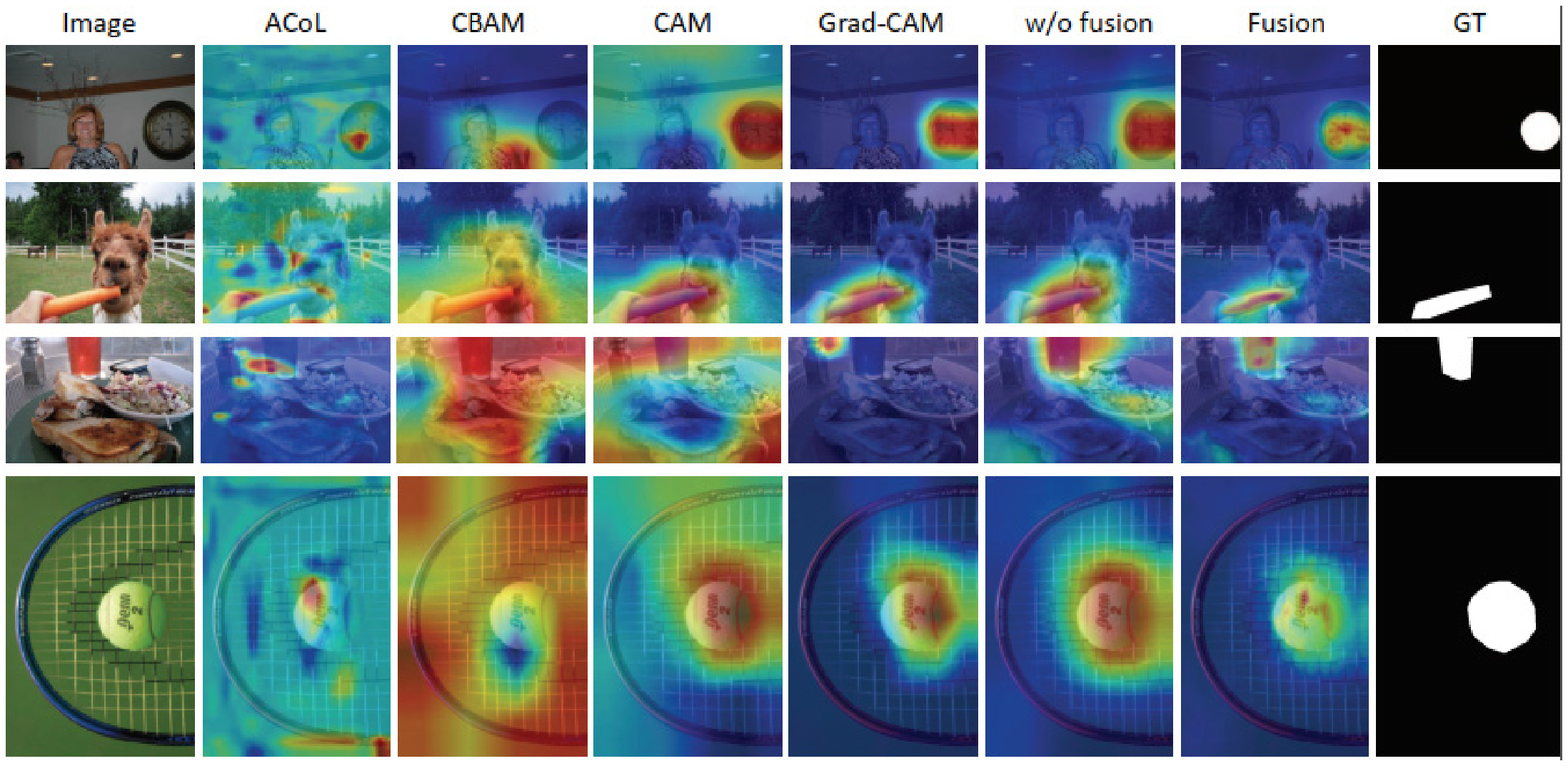}
\end{center}
   \caption{The class activation maps generated by the existing methods and the proposed method (``w/o Fusion'' and ``Fusion'').}\label{res_comparision_subjective}
\end{figure*}

The mIoU values on COCO 2017 dataset are shown in Table \ref{res_com}. ``Ours'' and ``Our w/o fusion'' are the results of our method using and without using the proposed multiple-layer based CAM generation method respectively. It is seen that using our multiple-layer based CAM generation method can obtain better results (15.41 vs 12.85). Furthermore, the mIoU values on COCO 2017 dataset are small, which is caused by the complicated backgrounds, the multiple class of objects and the multiple instances of each class in the images of COCO 2017 dataset. The mIoU value of each class by our method can be found in Fig. \ref{coco_res2}.
\begin{table*}
\small
\caption{The mIoU values by the existing methods and our method. The best backbone networks for different methods are used. PASCAL VOC 2012 and COCO 2017 dataset are considered.}
\begin{center}
%\resizebox{\textwidth}{30mm}{
\begin{tabular}{ccccc} %l(left)居左显示 r(right)居右显示 c居中显示
\hline
Method&Network&PASCAL-val& PASCAL-train & COCO-val\\ \hline
%CAM&ResNet50&21.70 & 25.49&\\ \hline
CAM \cite{zhou2016learning}&ResNet101&24.54 & 26.86&10.32\\ \hline
%Grad-CAM&ResNet50&25.49 &27.72 &\\ \hline
Grad-CAM \cite{selvaraju2017grad}&ResNet101&27.94 &27.94 &11.82\\ \hline
%Our w/o fuse&ResNet50&27.19 &29.53 & \\ \hline
ACoL \cite{zhang2018adversarial}&VGG19&19.42 & 19.41 &9.42\\ \hline
CBAM \cite{woo2018cbam}&ResNet50&17.63 & 21.08&9.67\\ \hline
Our w/o fusion&ResNet101&27.25 &29.39 &12.85\\ \hline
%Our&ResNet50&28.40 &29.72 &\\ \hline
Our&ResNet101&33.43 &34.43 &15.41\\ \hline
\end{tabular}
\end{center}
\label{res_com}
\end{table*}

\subsection{Comparisons with More Existing Methods}
We finally compare our method with several existing methods such as CAM \cite{zhou2016learning}\footnote{https://github.com/metalbubble/CAM}, Grad-CAM \cite{selvaraju2017grad}, ACoL \cite{zhang2018adversarial}\footnote{https://github.com/xiaomengyc/ACoL} and CBAM \cite{woo2018cbam} \footnote{https://github.com/Jongchan/attention-module}. The codes publicly released by the authors are used. Since the generation models on PASCAL VOC 2012 and COCO 2017 dataset are not provided in these source codes, we train the models by the two datasets. For CAM, we use ResNet101 as the backbone network. For ACoL, we use VGG19 (VGG16 is suggested in the released code) that can obtain better results as the backbone network. For CBAM, we use ResNet50 (obtains better results than ResNet101 in our experiments) as the backbone network.

Some subjective results are displayed in Fig. \ref{res_comparision_subjective}, where the original images, the activation maps by the existing methods and our method (``Fusion'' and ``w/o Fusion'' are our methods using and without using the multiple-layer based CAM generation) are displayed. Groundtruth (``GT'') is also displayed. It is seen that out method obtains better maps than the existing methods.

The mIoU values by the comparison methods and our method are displayed in Table \ref{res_com}. The detailed mIoU values by these methods on COCO 2017 dataset can be found in Fig. \ref{coco_res2}. It is interesting to see that the results by ACoL and CBAM are smaller than CAM and Grad-CAM methods. The reason is that ACoL is based on erasing strategy which performs well when the object is large, and the image contains single class of object. But when the objects are small, and the images contain multiple classes of objects as usually appeared in PASCAL VOC 2012 and COCO 2017 datasets, the objects are highly likely to be erased in the first stage, and second most activation regions by the second stage are usually the regions of different class of objects, which reduces the mIoU values. For CBAM, the attention model is used to enhance the convolution features. It focuses on attention regions that may belong to different classes. It is also seen that the proposed method outperforms the existing methods well, since our method selects representative classes for each class that can avoid the interferences of the classes, and generate better class activation maps.

\section{Conclusion}
This paper proposes a new class activation map generation method by selecting representative classes, with the aim of fusing the complementary regions activated by different classes. A clustering based class selection method is first proposed. The classification based similarity matrix construction and the improved k-means clustering method are proposed for the selection. Then, we propose a multi-layer based CAM generation method. The multiple binary classification based CAM generation with a simple fusion method is finally introduced to generate the activation map. Experimental results show that the proposed method can improve the basic Grad-CAM method obviously. Moreover, small CNN networks also work well on our method due to our multiple binary classification based strategy.

{\small
\bibliographystyle{ieee}
\bibliography{egbib}
}

\end{document}